\documentclass[letterpaper, 10pt, conference]{ieeeconf}
\IEEEoverridecommandlockouts
\usepackage[acronym]{glossaries}
\usepackage{amsmath,amssymb,amsfonts}
\usepackage{algorithmic}
\usepackage{graphicx}
\usepackage{subcaption}
\usepackage{caption}
\usepackage{svg}
\usepackage{textcomp}
\usepackage{xcolor}
\let\labelindent\relax
\usepackage{enumitem}
\usepackage{mathrsfs}

\usepackage{placeins}
\usepackage{todonotes}
\usepackage{booktabs}
\usepackage{multirow}
\usepackage{fontawesome5}
\usepackage{tikz}
\usepackage{makecell}
\usepackage{cite}
\DeclareMathOperator*{\argmin}{arg\,min}
\newcommand{\mymodel}{MatchInformer }

\title{Don't double it: Efficient Agent Prediction in Occlusions
}
\author{Anna Rothenhäusler\textsuperscript{1}, Markus Mazzola\textsuperscript{2}, Andreas Look\textsuperscript{2}, Raghu Rajan\textsuperscript{1}, Joschka Bödecker\textsuperscript{1}}
\begin{document}
\maketitle

\thispagestyle{empty}
\pagestyle{empty}

\begingroup
\renewcommand\thefootnote{\textsuperscript{\arabic{footnote}}}
\raggedright
\footnotetext[1]{Department of Computer Science, University of Freiburg, Germany. \texttt{\{rothenha, rajanr, jboedeck\}@cs.uni-freiburg.de}}
\footnotetext[2]{Bosch Center for Artificial Intelligence, Germany. \texttt{\{markus.mazzola, andreas.look\}@de.bosch.com}}
\endgroup



\begin{abstract}
Occluded traffic agents pose a significant challenge for autonomous vehicles, as hidden pedestrians or vehicles can appear unexpectedly, yet this problem remains understudied.
Existing learning-based methods, while capable of inferring the presence of hidden agents, often produce redundant occupancy predictions where a single agent is identified multiple times. This issue complicates downstream planning and increases computational load. 
To address this, we introduce MatchInformer, a novel transformer-based approach that builds on the state-of-the-art SceneInformer architecture. Our method improves upon prior work by integrating Hungarian Matching, a state-of-the-art object matching algorithm from object detection, into the training process to enforce a one-to-one correspondence between predictions and ground truth, thereby reducing redundancy. We further refine trajectory forecasts by decoupling an agent's heading from its motion, a strategy that improves the accuracy and interpretability of predicted paths. To better handle class imbalances, we propose using the Matthews Correlation Coefficient (MCC) to evaluate occupancy predictions. By considering all entries in the confusion matrix, MCC provides a robust measure even in sparse or imbalanced scenarios. Experiments on the Waymo Open Motion Dataset demonstrate that our approach improves reasoning about occluded regions and produces more accurate trajectory forecasts than prior methods.
\end{abstract}


\section{Introduction}

Reasoning about occluded traffic agents is a major challenge for autonomous vehicles. Occlusions hide agents such as pedestrians behind parked cars or vehicles approaching blind intersections. Human drivers excel at this by drawing on experience and transferring knowledge between situations, anticipating potential hazards from other agents. For example, a cyclist will often slow down when riding parallel to parked cars, anticipating that a car door might open unexpectedly. Early approaches used worst-case assumptions, where the system would assume the most dangerous possible outcome, such as an infinitely long vehicle emerging from an occlusion at maximum speed~\cite{worst_case_1, worst_case_2, worst_case_3}. Although this ensures safety by planning for the highest possible risk, it can lead to overly cautious and inefficient driving, causing deadlocks in dense traffic, or even creating dangerous situations by making the autonomous vehicle itself unpredictable \cite{worst_case_critic, worst_case_1}. 

\begin{figure}[t]
    \centering
    \begin{subfigure}{0.98\linewidth}
        \centering
        \includegraphics[scale=0.6, trim=7cm 0.1cm 5cm 25cm, clip]{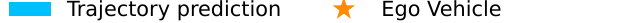}
    \end{subfigure}
    \begin{subfigure}{0.43\linewidth}  
        \centering
        \includegraphics[height=4.0cm, keepaspectratio, trim=0cm 13cm 7cm 0.2cm, clip]{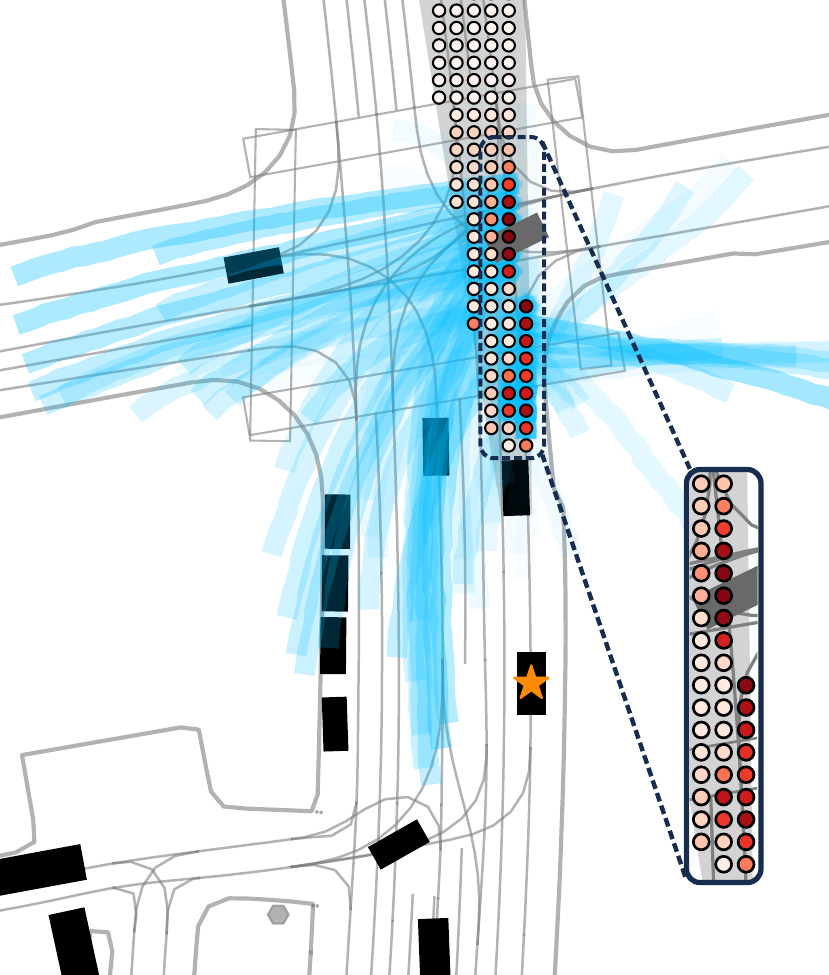}
        \caption{SceneInformer (prior art)}
        \label{fig:traj_original}
    \end{subfigure}
    \hfill
    \begin{subfigure}{0.43\linewidth}
        \centering
        \includegraphics[height=4.0cm, keepaspectratio, trim=0cm 13cm 7cm 0.2cm, clip]{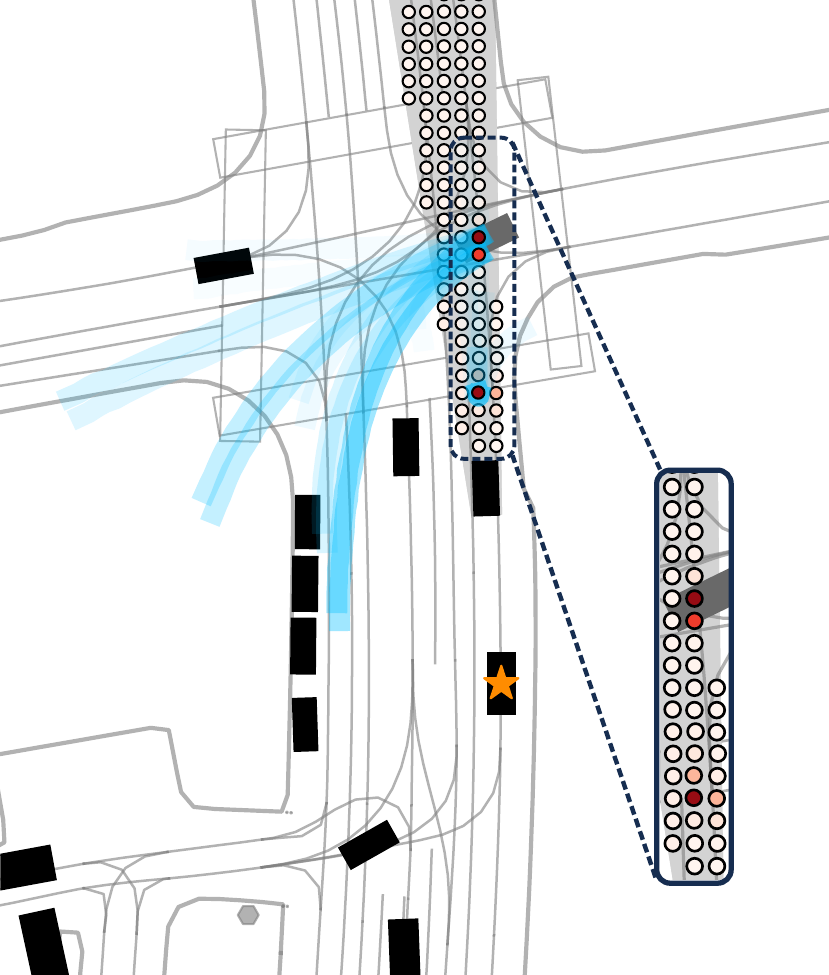}
        \caption{MatchInformer (ours)}
        \label{fig:traj_mymodel}
    \end{subfigure}
    \hfill       
    \begin{subfigure}{0.10\linewidth}
        \centering
        \raisebox{0.05\height}{\includegraphics[width=0.72\linewidth, trim=20.0cm 7cm 0.0cm 0.1cm, clip]{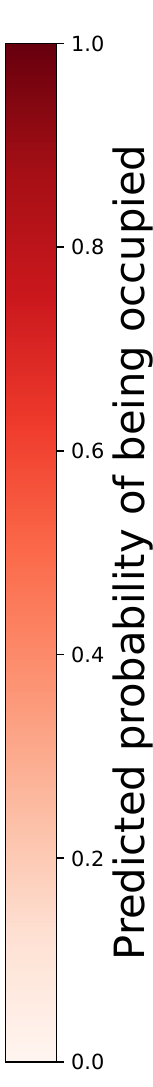}}
    \end{subfigure}

    \caption{We visualize our model's predictions in an occlusion scenario, contrasting them with those of the prior work, SceneInformer. As shown, SceneInformer (a) predicts a high density of occupied points, which generates numerous candidate trajectories and creates a cluttered scene for the planner. In stark contrast, our model (b) produces a much sparser and more refined prediction, demonstrating its ability to reduce redundant occupancy forecasts. In these visualizations, occupancy probability is indicated by a gradient from white (0) to deep red (1), predicted trajectories are shown in blue, and only trajectories for points with a probability greater than 0.5 are displayed.}
    \label{fig:traj_comparison}
\end{figure}

Recent learning-based methods discretize occlusions into a grid and predict the probability of each cell being occupied~\cite{OGM_1, PaS_ref1, PaS_ref2}. Extensions, such as SceneInformer~\cite{SceneInformer}, set anchor points on each grid cell, treating these points as potential agents for which they generate trajectory predictions. However, reasoning about occlusions remains challenging: redundant occupancy predictions often produce a large number of candidate trajectories, which in turn increase the computational load on planning modules, as can be seen in Fig.\,\ref{fig:traj_original}. Addressing this redundancy is therefore a central motivation for the model proposed in this work. As shown in Fig.\,\ref{fig:traj_mymodel}, our approach reduces false positive occupancy predictions, thereby preventing the generation of redundant trajectories and lowering the number of candidates that must be considered by the planner.

Building on the work of SceneInformer~\cite{SceneInformer}, we propose a new training methodology incorporating Hungarian Matching. Prior to calculating the loss, this method optimally assigns the predicted agents to the ground-truth agents. Without this method, the model would only be rewarded for a prediction that exactly matches the ground-truth position, ignoring any close but not perfectly aligned predictions. Since determining the exact position of agents in occluded areas is inherently uncertain, this strict matching could unfairly penalize the model for reasonable predictions. The model takes the observed agent trajectories as input and predicts the occupancy of occluded regions, as well as the future trajectories of both visible and occluded agents. Instead of absolute positions, our model predicts the heading and relative trajectory of each agent, rotating the latter by the heading to represent motion in a local, agent-centered frame while preserving global orientation. We adapted the architecture to include a new prediction head, which distinguishes between different classes (e.g., cars, pedestrians, and bicycles). This is a change from the previous binary classification of occupied versus unoccupied space. Training and evaluation are performed on the Waymo Open Motion Dataset (WOMD) \cite{womd}. At each history timestep, ray-casting is used to estimate the ego vehicle’s field of view by projecting rays from the origin of the ego sensor to detect visible areas and identify occluded regions. This procedure allows the model to determine which parts of the scene are currently observable and which are hidden behind obstacles, enabling occlusion-aware occupancy and trajectory predictions. To evaluate classification accuracy, we adopt the Matthews Correlation Coefficient (MCC)\,\cite{MCC}. The MCC is more robust to class imbalance than standard metrics such as accuracy or the F1-score, because it considers all possible outcomes — both correct and incorrect predictions for positive and negative classes — providing a balanced assessment of overall model performance. We compare our experimental results with SceneInformer\,\cite{SceneInformer}, which, to the best of our knowledge, is the only recent open-source method for occluded trajectory and occupancy prediction that enables reproducible evaluation. Our approach demonstrates improvements in both trajectory and occupancy prediction.

Our contributions are manyfold: 
\begin{itemize}
    \item We address the challenge of redundant predictions by applying Hungarian Matching prior to loss computation.
    \item Refine trajectory forecasts by decoupling the heading from motion prediction and rotating the relative trajectories according to the agent’s heading.
    \item We adapt the existing prediction head to perform multiclass classification, distinguishing between cars, pedestrians, and bicycles.
    \item Demonstrate state-of-the-art performance in trajectory prediction and occupancy prediction in occlusions, and introduce MCC in this context as a measure for evaluating imbalanced occupancy predictions.
\end{itemize}

\section{Related Work}

\textbf{Occupancy Prediction of Occlusions} is essential for anticipating hidden agents and ensuring safe autonomous navigation. Rather than relying solely on hardware sensors like lidar or cameras, we can use people as "sensors" or sources of additional information. This concept, often called People as Sensors (PaS), leverages the fact that human actions and behaviours can provide crucial data that complement hardware-based sensing \cite{PaS_ref1, PaS_ref2}. For instance, the sudden deceleration of a car approaching a crosswalk often signals a pedestrian's presence before they are even in view. Building on this concept, Itkina et al. \cite{PaS} recognize that interpreting these social cues is an inherently multimodal and uncertain problem. The deceleration of a vehicle directly ahead provides a reasonable inference of an obstacle or event in its forward path. However, this inference is by nature multimodal because a single observed behavior can arise from a variety of different underlying situations. For instance, while the deceleration may signal the presence of a pedestrian at a crosswalk, it could just as well indicate a different scenario, such as traffic congestion ahead or the unexpected detection of an available parking space. To address this, they employ a conditional variational autoencoder (CVAE) to predict plausible fixed-size occupancy maps in front of drivers, based solely on agent motion, without relying on explicit road layout information. Building on this approach, Christianos et al. \cite{Bi-Level} introduce an additional CVAE to predict probable future trajectories emerging from the occlusion based on the predicted occupancy map. The limitations of this approach are that it relies on a fixed-size occlusion inference that only applies to areas behind hidden agents. Furthermore, the model does not account for interactions between agents, and while predicted overlapping maps are fused, a single prediction is based on the behavior of only one agent.
Lange et al., the authors of SceneInformer~\cite{SceneInformer}, address these limitations by introducing a transformer-based encoder-decoder model that predicts both occupancy probabilities and future trajectories for a set of anchor points within an occluded region. However, because their one-to-one matching, pairing predicted points to ground-truth agents, is performed solely based on spatial position, the model can produce multiple false positives clustered around ground truth objects. This behaviour arises because the training loss, specifically the weighted cross-entropy, is designed to penalize missed true positives more heavily than redundant detections. In practice, anchor points that correspond to truly occupied positions are assigned a weight of 50, ensuring that the model focuses on correctly predicting these sparse positive points. Such weighting is necessary to mitigate the severe class imbalance, as unoccupied anchor points vastly outnumber those corresponding to actual agents. As a consequence, the model is not sufficiently penalized for duplicate detections, leading to ambiguous supervision and redundant predictions near the same ground truth. Since distinct trajectory modes are predicted for each anchor point, the total number of possible trajectories can become very large, as illustrated in Figure\,\ref{fig:traj_comparison}. To address this, redundancy reduction strategies commonly used in object detection can be applied.


\textbf{Redundancy Reduction} in object detection is a crucial problem. Many object detection algorithms \cite{SSD, R-CNN, YOLO} use non-maximum suppression (NMS) to reduce the number of false positives. NMS filters out redundant overlapping detections. The most commonly used metric for the selection process is the Intersection over Union (IoU), although other metrics, such as the Euclidean distance, are also employed. Among overlapping candidates, only those with the highest confidence and above the designated threshold are kept.
DETR \cite{DETR}, a transformer encoder-decoder-based end-to-end object detection framework, incorporates the one-to-one matching strategy \textit{Hungarian Matching} within the training and makes an additional postprocessing step unnecessary. Hungarian Matching \cite{HungarianMatching} solves the optimal assignment problem, finding the best one-to-one assignment between ground truth and predicted boxes, teaching the model not to duplicate predictions for the same object.

\par

We propose a transformer-based encoder-decoder model that incorporates Hungarian Matching prior to loss computation. To the best of our knowledge, this is the first application of Hungarian Matching for autonomous driving occlusion-aware agent prediction tasks, enabling optimal assignment of predicted agents to ground-truth targets even in multimodal scenarios where occlusions or limited visual information make exact positions uncertain.

\begin{figure*}[t]
  \centering
  \includegraphics[width=0.7\textwidth, trim=0cm 6.7cm 0cm 2.5cm, clip]{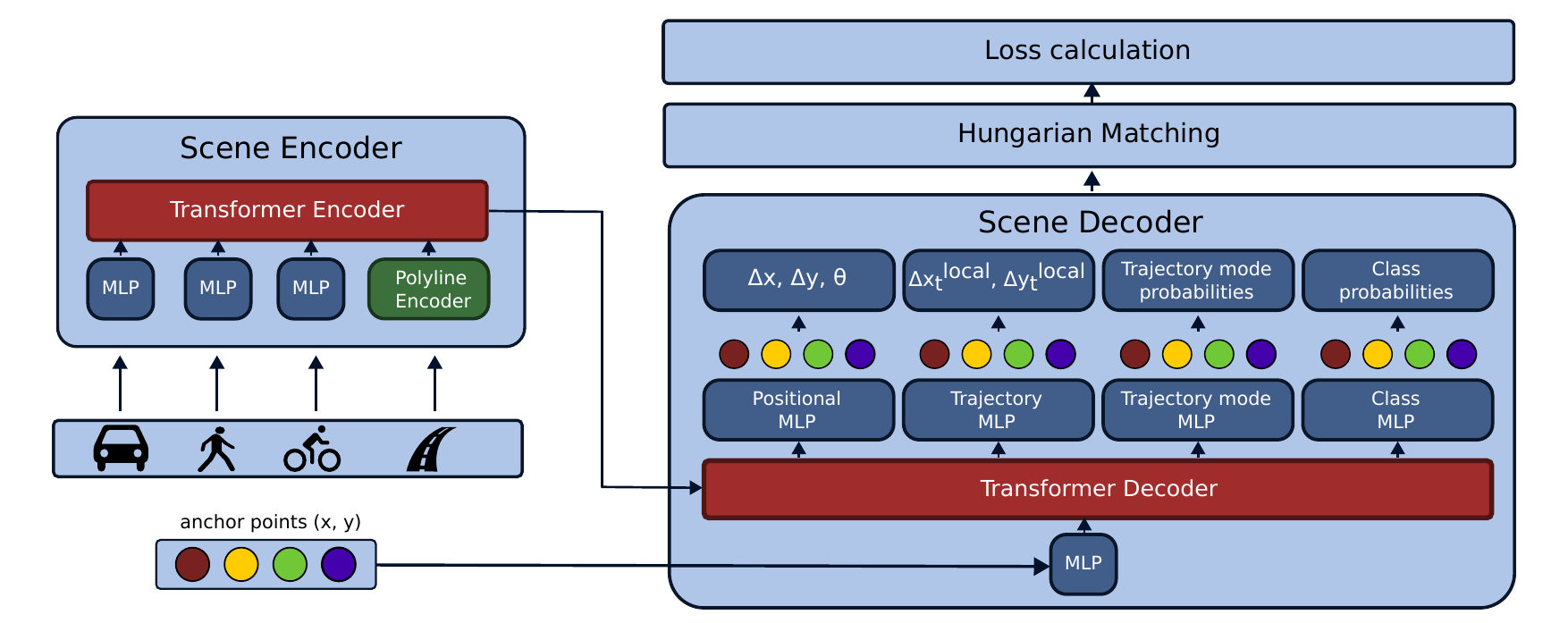}
  \caption{Overview of the MatchInformer training process: Key improvements over SceneInformer include Hungarian Matching before loss computation, a positional MLP that allows additional movement of the anchor points and predicts heading angles, and a class MLP that predicts occupancy and assigns each point to a class (car, pedestrian, or bicycle).}
  \label{fig:MatchFormer}
\end{figure*}
\section{\mymodel}

We propose MatchInformer, a transformer encoder-decoder-based combination of the SceneInformer architecture as well as the object detection model DETR. The whole training process is illustrated in Figure\,\ref{fig:MatchFormer}. The encoder structure is adopted from SceneInformer~\cite{SceneInformer} and operates in a feature-based manner, encoding observed trajectories of other agents using MLPs. A separate MLP is employed for each of the other agent types including cars, pedestrians, and bicycles. In addition, a polyline encoder encodes the road layout. In line with the DETR paradigm,\cite{DETR}, anchor points are introduced as reference $(x,y)$ locations that serve as starting positions for the decoder’s predictions. Following \cite{SceneInformer}, we use a combined set of anchor points derived from two sources: the most recent positions of observed agents and a grid at 1.5-meter intervals within occluded areas to represent hidden agents. The decoder uses an MLP as a position encoding function \cite{SceneInformer, AnchorDETR} for these anchor points. The transformer decoder is followed by four prediction heads, responsible for estimating the corresponding class, positional displacement, future trajectories and associated trajectory mode probabilities. Here, a “mode” represents one of several possible future paths an agent might take, reflecting the inherent uncertainty and multimodality in predicting agent behavior. Unlike SceneInformer \cite{SceneInformer}, which predicts only a probability distribution over occupied versus unoccupied points, our classification head outputs a probability distribution over all possible classes: car, bicycle, pedestrian, and \textit{no class}. We follow current literature \cite{DETR} in predicting a “no class” label, which allows points to be classified as background. This is particularly important in occluded regions, where the number of potential points exceeds the actual number of agents, enabling the model to effectively distinguish relevant agents from background. In contrast to SceneInformer, we introduce an additional positional head that predicts potential shifts in the \(x\) and \(y\) directions, denoted by \(\Delta x\) and \(\Delta y\), respectively.
Furthermore, the positional head predicts the heading angle $\theta$ of each anchor point, reflecting the orientation of the associated traffic participant. This explicit heading prediction enables the model to condition trajectory forecasting on orientation, thereby focusing solely on the relative positional displacement with respect to the current position. By doing so, the model learns relative trajectory modes anchored to each heading direction, such as moving straight, turning left or right, or adjusting speed. Incorporating heading information in this manner eliminates the need to implicitly encode orientation through coordinate transformations and helps to decouple rotational information from translational motion, facilitating a cleaner separation of movement patterns in the trajectory prediction process. Instead of predicting the heading angle $\theta$ directly, the model predicts $\sin(\theta)$ and $\cos(\theta)$ to avoid angle discontinuities near $\pm\pi$, ensuring numerical stability during training. In addition, these values are normalized to lie on the unit circle, which guarantees consistency between the predicted components. 
Following previous work,\cite{SceneInformer, MotionTransformer}, two heads are used for trajectory prediction: one estimates the probabilities of the $M$ predicted trajectory modes, and the other predicts the local \(x\) and \(y\) displacements for each mode and each future timestep $t$. In contrast to this work, these local displacements are defined in the vehicle coordinate frame — that is, as if observed from the vehicle’s own perspective — and must be rotated by the heading angle $\theta$ to obtain displacements in the global coordinate frame, corresponding to the trajectory observed from an external point of view
\begin{equation}
\begin{pmatrix}
\Delta x_{t}^{\text{global}} \\
\Delta y_{t}^{\text{global}}
\end{pmatrix}
=
\begin{pmatrix}
\cos(\theta) & -\sin(\theta) \\
\sin(\theta) & \cos(\theta)
\end{pmatrix}
\begin{pmatrix}
\Delta x_{t}^{\text{local}} \\
\Delta y_{t}^{\text{local}}
\end{pmatrix}.
\end{equation}
All prediction heads are implemented using MLPs.

\textbf{Hungarian Matching} is a common technique that computes an optimal one-to-one assignment between two sets based on a given cost matrix~\cite{HungarianMatching}. Hungarian Matching assigns predictions to ground-truth instances through an optimal one-to-one mapping that minimizes a cost matrix by solving the linear assignment problem \cite{LinearAssProblem}. It produces unambiguous labels, removes duplicate assignments, and increases training stability. It also effectively handles cases where the number of predictions differs from the number of ground-truth agents. Let $G$ be the set of ground-truth object indices and $N$ the set of prediction indices (i.e. the number of anchor points input to the decoder), with $|G| < |N|$. Hungarian Matching finds an optimal one-to-one assignment $\sigma : N \rightarrow G \cup \{\varnothing\}$ that minimizes the total cost. Each prediction $n \in N$ is either assigned to a ground-truth object $\sigma(n) = g \in G$ or to "no object" $\varnothing$, representing the absence of a ground-truth match. Based on a cost matrix that includes positional and classification costs, this assignment allows the model to reward predictions that are close to the ground truth, not just those that are perfectly positionally aligned. 
Let $\mathbf{C} \in \mathbb{R}^{(|G|+1) \times |N|}$ be the cost matrix, where $\mathbf{C}_{g,n}$ represents the cost of matching prediction $n \in N$ with ground-truth object $g\in G$, and the extra row $C_{\varnothing,n}$ corresponds to assigning prediction $n$ to ``no object.'' The total assignment cost is
\[
\text{Cost}(\sigma) = \sum_{n \in N} \mathbf{C}_{\sigma(n), n}.
\]

\begin{figure}[ht]
    \centering
    \begin{subfigure}{0.43\linewidth}
        \centering
        \includegraphics[width=0.9\linewidth, trim=2.7cm 18cm 9.7cm 2cm, clip]{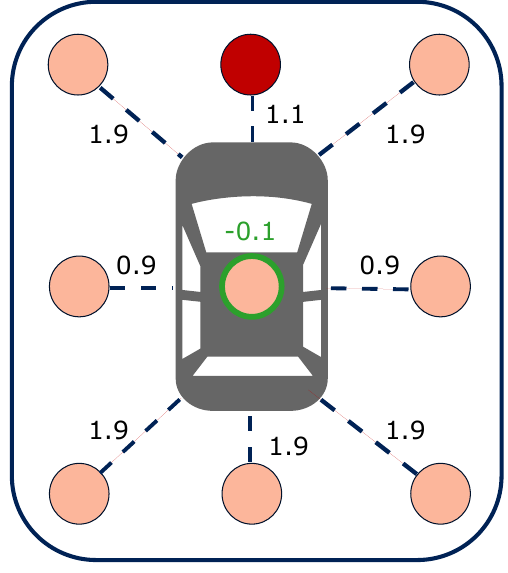}
        \caption{$\lambda_{\text{pos}} = 1.0, \lambda_{\text{class}} = 1.0$}
        \label{fig:low_cost}
    \end{subfigure}
    \hfill
    \begin{subfigure}{0.43\linewidth}
        \centering
        \includegraphics[width=0.9\linewidth, trim=6.2cm 17.5cm 6.2cm 2.5cm, clip]{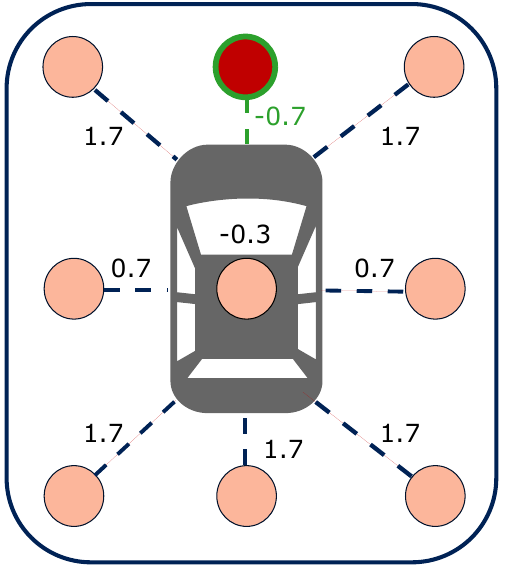}
        \caption{$\lambda_{\text{pos}} = 1.0, \lambda_{\text{class}} = 3.0$}
        \label{fig:high_cost}
    \end{subfigure}
    \begin{subfigure}{0.10\linewidth}
        \centering
        \raisebox{0.05\height}{\includegraphics[width=0.7\linewidth, trim=20.0cm 7cm 0.0cm 0.1cm, clip]{figures/legend/blue_greenred_bar_legend.pdf}}
    \end{subfigure}

    \caption{Hungarian Matching is performed between predicted grid points and either the ground-truth car or the “no class” label. The point highlighted with a green outline is matched to the car, all others are assigned to “no class.” In Fig.~\ref{fig:low_cost}, a lower class weight ($\lambda_{\text{class}} = 1.0$) matches the point directly on the car despite its low occupancy probability. In Fig.~\ref{fig:high_cost}, a higher weight ($\lambda_{\text{class}} = 3.0$) matches a point above the car, as higher occupancy outweighs distance. Edge numbers show the matching costs to the ground-truth car.}
    \label{fig:HungarianMatching_example}
\end{figure}

Hungarian Matching finds the optimal mapping
\[
\hat{\sigma} = \argmin_{\sigma: N \to G \cup \{\varnothing\}} \sum_{n \in N} \mathbf{C}_{\sigma(n), n},
\]
subject to the one-to-one constraint that each ground-truth object $g \in G$ is assigned to exactly one prediction $n \in N$, while predictions may be assigned to at most one object (or to ``no object'').
Each cost matrix entry $\mathbf{C}_{g,n}$, corresponding to assigning prediction $n \in N$ to ground-truth object $g \in G$, is defined as a weighted sum of the classification cost — derived from the predicted probability of the correct class — and the L2 distance between predicted and ground-truth positions.  
Let the cost of assigning prediction $n \in N$ to ground-truth object $g \in G$ be 
\begin{equation}
C_{g,n} = \lambda_{\text{pos}} \, \text{L2}(\mathbf{\hat{p}}_n, \mathbf{p}_g) 
- \lambda_{\text{class}} \, \hat{\mathbf{z}}_{n}^\top \mathbf{z}_g,
\end{equation}

where 
\(\mathbf{\hat{p}}_n = (\hat{x}_n, \hat{y}_n) \in \mathbb{R}^2\) is the predicted position of anchor point $n$, 
\(\mathbf{p}_g = (x_g, y_g)\in \mathbb{R}^2\) is the ground-truth position of agent $g$, 
\(\hat{\mathbf{z}}_n \in \mathbb{R}^4\) is the predicted class probability vector, 
and $\mathbf{z}_{g} \in \{0, 1\}^4$ denotes the one-hot encoding of the ground-truth class.
\(\lambda_{\text{pos}}, \lambda_{\text{class}} \in \mathbb{R}\) are weighting factors. The weighting factors are introduced to allow a certain tolerance in distance, specifying a range within which predicted points can be considered correctly matched to ground-truth positions. Without these weights, the positional error would dominate the matching process, overwhelming the influence of the class probabilities as shown in Fig.~\ref{fig:HungarianMatching_example}.


\begin{figure*}[ht]
    \vspace{0.3cm}
    \centering
    \hfill
    \begin{subfigure}{0.31\textwidth}
        \centering
        \includegraphics[width=\linewidth]{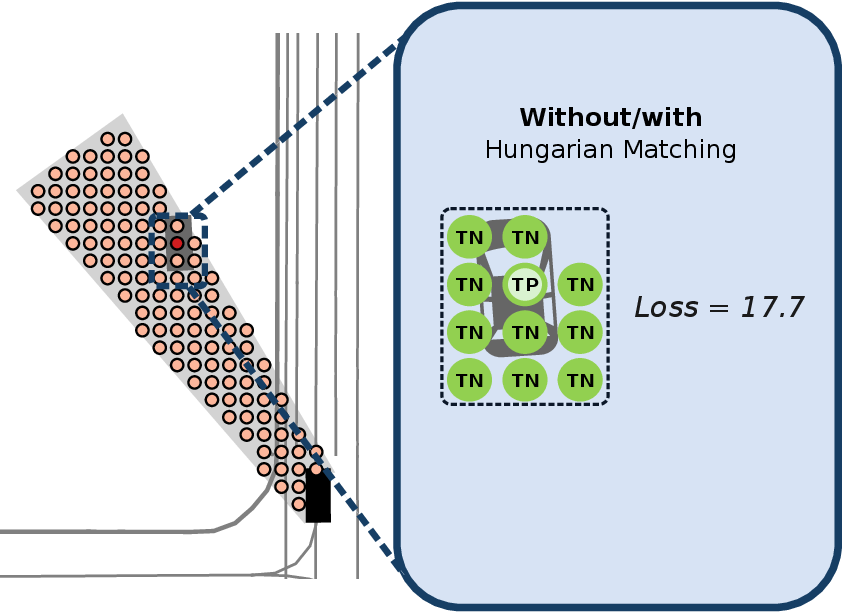}
        \caption{perfect prediction}
    \end{subfigure}
    \hfill
    \begin{subfigure}{0.31\textwidth}
        \centering
        \includegraphics[width=\linewidth]{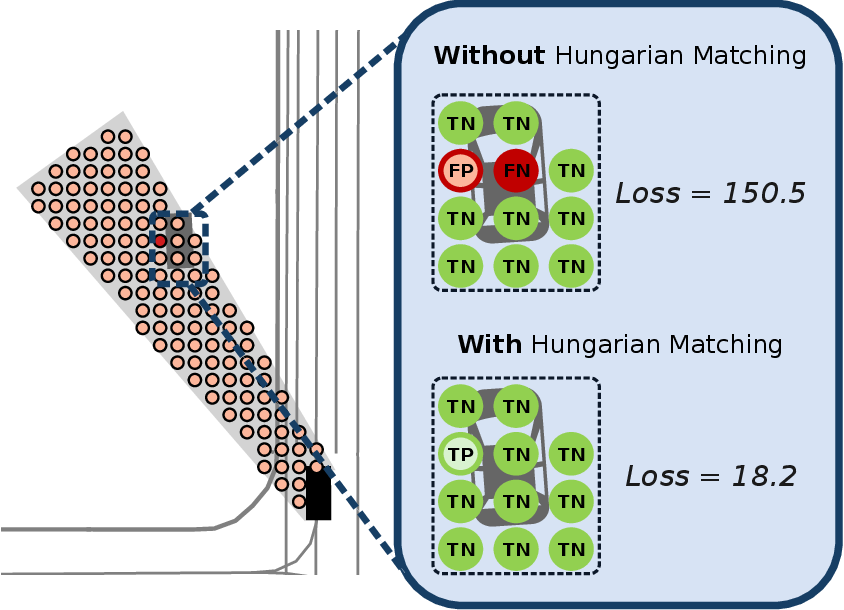}
        \caption{offset prediction}
        \label{fig:offset_point}
    \end{subfigure}
    \hfill
    \begin{subfigure}{0.31\textwidth}
        \centering
        \includegraphics[width=\linewidth]{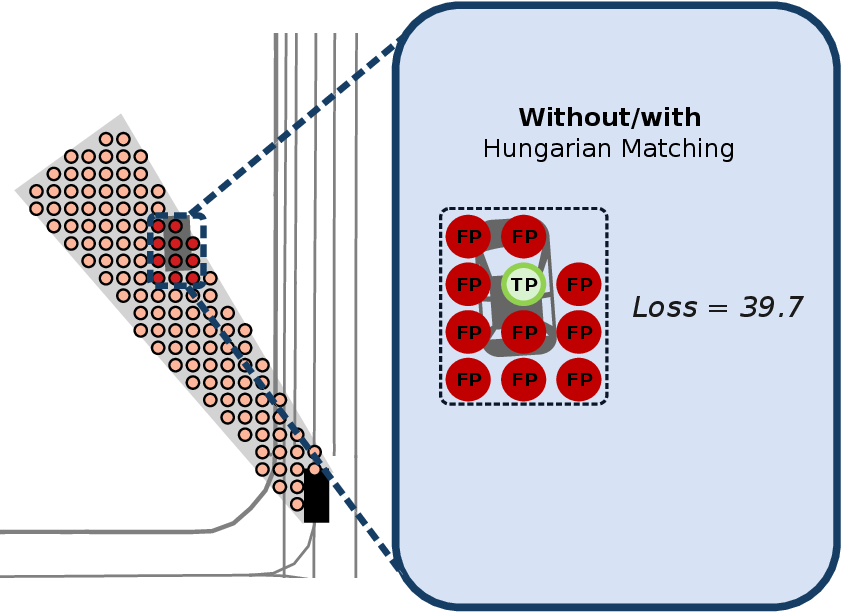}
        \caption{redundant predictions}
        \label{fig:multiple_wrong}
    \end{subfigure}
    \hfill

   \caption{Weighted cross-entropy loss for three prediction scenarios, shown without Hungarian Matching (SceneInformer) and with Hungarian Matching (MatchInformer, our model). Since each point represents one agent, a car is matched to a single point. In case (a), the exact car position is predicted correctly. In case (b), a neighboring point is predicted as occupied, while the exact-position point has a low predicted probability. In case (c), both the exact-position point and surrounding points are predicted with high occupancy probability. Here, true positives (TP) are correctly predicted occupied points, false positives (FP) are points predicted as occupied but actually free, false negatives (FN) are missed occupied points, and true negatives (TN) are correctly predicted free points. The loss differs only in (b): without Hungarian Matching, slightly offset predictions result in both a false negative and a false positive, which significantly increases the loss. With Hungarian Matching, the offset prediction is paired with the ground-truth object, incurring only a small additional positional cost.}
    \label{fig:HungarianMatching}
\end{figure*}
In Fig.~\ref{fig:HungarianMatching}, the impact of applying Hungarian Matching before loss calculation can be clearly observed. Without Hungarian Matching, the loss increases when a predicted point does not exactly match the correct position, even if it is close to the ground truth (see Fig.~\ref{fig:offset_point}). In fact, predictions that do not exactly match the ground-truth position can incur a higher loss than predicting all points in the region as unoccupied. This can unintentionally encourage the model to predict multiple occupied points in an attempt to hit the exact target, rather than risking missing it entirely. By contrast, when using Hungarian Matching, a point that deviates from the ground-truth position can still count as a true positive, incurring only a minor penalty for its positional offset. This approach encourages the model to make fewer positive predictions within the region of interest, rather than producing additional positive predictions to compensate for potential positional mismatches.

The \textbf{Training Loss} consists of three terms for anchor points matched to a ground truth object. 
The first term is the classification loss $\mathcal{L^{\text{class}}}$, defined as the cross-entropy (CE) between the predicted logits vector $\hat{\mathbf{z}}_{n}$ and the corresponding ground-truth one-hot class vector $\mathbf{z}_{\sigma(n)}$ 
\begin{equation}
    \mathcal{L}_{n}^{\text{class}} = \mathrm{CE}(\hat{\mathbf{z}}_{n}, \mathbf{z}_{\sigma(n)}).
\end{equation}
If the point $n$ is matched to "no class", the class loss becomes $\mathcal{L}_{n}^{\text{class}} = \mathrm{CE}(\hat{\mathbf{z}}_{n}, \varnothing)$.

The second term is the positional loss $\mathcal{L^{\text{pos}}}$, consisting of two components. First, a mean squared error (MSE) is computed between the predicted anchor point position $\hat{\mathbf{p}}_{n}$ and the ground truth position of the matched point $\mathbf{p}_{\sigma(n)}$.
Second, cosine similarity serves as the loss function for the prediction of the heading angle $\hat{\theta}_n$. The angle is represented as a 2D unit vector $\hat{\mathbf{u}}_{n} = (\cos(\hat{\theta}_n), \sin(\hat{\theta}_n)) \in \mathbb{R}^2$ which is compared to the ground truth orientation vector $\mathbf{u}_{\sigma(n)} = (\cos(\theta_{\sigma(n)}), \sin(\theta_{\sigma(n)})) \in \mathbb{R}^2$. This approach penalizes angular differences between the two directions, making it well-suited for rotation-invariant orientation estimation. The total positional loss is computed as
\begin{equation}
    \mathcal{L}_{n}^{\text{pos}} = \mathrm{MSE}(\mathbf{\hat{p}}_{n}, \mathbf{p}_{\sigma(n)}) + (1 - \hat{\mathbf{u}}_{n}^\top \mathbf{u}_{\sigma(n)}).
\end{equation}
The third term corresponds to the trajectory prediction loss, $\mathcal{L}^{\text{traj}}$, which consists of two components. 
Let $m_n^* \in \{1, \dots, M\}$ denote the index of the mode closest to the ground-truth trajectory for prediction $n$. 
We define $\mathbf{\hat{y}}_n \in \mathbb{R}^M$ as the predicted probability vector over the $M$ modes, and $\mathbf{y}_{\sigma(n)} \in \{0,1\}^M$ as the corresponding one-hot vector indicating the closest mode $m_n^*$, i.e., $\mathbf{y}_{\sigma(n),m_n^*} = 1$. This effectively implements a winner-takes-all (WTA) strategy, where only the mode closest to the ground truth contributes to the loss. 

The first component of $\mathcal{L}^{\text{traj}}$ is the cross-entropy (CE) loss between $\mathbf{\hat{y}}_n$ and $\mathbf{y}_{\sigma(n)}$, encouraging the model to assign high probability to the mode closest to the ground truth. 
The second component captures the trajectory regression loss, computed as the mean squared error (MSE) between the predicted positions $\mathbf{\hat{x}}_{n, t}^{m_{n}^*}$ of the closest mode $m_n^*$ and the ground-truth positions $\mathbf{x}_{\sigma(n),t}$ at each future timestep $t \in T$.
\begin{equation}
    \mathcal{L}_{n}^{\text{traj}} = \mathrm{CE}(\mathbf{\hat{y}}_{n}, \mathbf{y}_{\sigma(n)}) + \frac{1}{T}\sum_{t \in T} \mathrm{MSE}(\mathbf{\hat{x}}_{n, t}^{m^*_n}, \mathbf{x}_{\sigma(n), t}),
\end{equation}
where 
\begin{equation}
    m^*_n = \argmin_{m \in M} \frac{1}{T}\sum_{t \in T} \mathrm{MSE}(\mathbf{\hat{x}}_{n, t}^{m}, \mathbf{x}_{\sigma(n), t}).
\end{equation}
For anchor points not matched to any ground truth object, the model minimizes the cross-entropy loss for the "no class" ($\varnothing$) category.
The total loss function is a weighted combination of the previously described components of the loss:
\begin{align}
\mathcal{L} = {} & \sum_{n \,:\, \sigma(n) \neq \emptyset} \Big[
    \omega_{1}\mathcal{L}_{n}^{\text{class}} + \omega_{2}\mathcal{L}_{n}^{\text{pos}} + \omega_{3}\mathcal{L}_{n}^{\text{traj}} \Big] \nonumber \\
& + \sum_{n \,:\, \sigma(n) = \emptyset} \omega_{1}\mathcal{L}_{n}^{\text{class}}.
\label{eq:loss_equation}
\end{align}

\FloatBarrier

\section{Experiments}
\subsection{Dataset}
Our model is trained and validated on the Waymo Open Motion Dataset (WOMD) \cite{womd}. Following \cite{SceneInformer}, occlusions are incorporated into the bird’s-eye view (BEV) by ray-casting from the ego vehicle. For each scenario, a 1-second history is used to predict the following 4 seconds, with predictions made every 0.1\,s.

\subsection{Metrics}
The \textbf{Classification Accuracy} in SceneInformer~\cite{SceneInformer} is reported using sensitivity and specificity. However, these metrics are known to have limited practical value in the presence of a class imbalance~\cite{MCC}. Specifically, while high sensitivity indicates that most positive samples are correctly identified, it does not account for the proportion of predicted positives that are actually correct. Evaluating the precision of SceneInformer at the thresholds corresponding to high sensitivity and specificity results in an average precision of 0.0184 across all occlusion levels. In contrast, the Matthew Correlation Coefficient (MCC) is widely recognized to be more robust to class imbalance as it considers all elements of the confusion matrix. However, MCC is not entirely independent of the class distribution\,\cite{MCC, MCC_critic}. It ranges from -1 to +1, while -1 indicates completely inverse prediction, 0 corresponds to random guessing, and +1 signifies perfect prediction. According to the authors of \cite{MCC}, MCC is unique in that a high MCC score reflects a strong performance on all four fundamental metrics: sensitivity $\frac{TP}{TP+FN}$, specificity $\frac{TN}{TN+FP}$, precision $\frac{TP}{TP+FP}$ and negative predictive value $\frac{TN}{TN+FN}$. In contrast, the F1-score $\frac{2\cdot \text{precision} \cdot \text{sensitivity}}{\text{precision} + \text{sensitivity}}$ ignores true negatives (TN) and can therefore be misleadingly low when correctly identifying the negative class is also important. For example, in Figure\,\ref{fig:multiple_wrong}, a model that successfully identifies a rare positive case might still only achieve an F1-score of 0.18. In contrast, relying on sensitivity and specificity in this example could be deceptive, as their high values of 1.0 and 0.92, respectively, would give the false impression of a highly accurate model. The MCC provides a more accurate and balanced evaluation, resulting in a score of 0.28 for the same model. The MCC is a more suitable evaluation metric because it accounts for the true negatives, which are the overwhelming majority of points in the dataset, and thus provides a more reliable measure of the model's overall performance. For this reason, MCC is employed here as a classification evaluation metric, reporting it only for the occluded area.
The MCC is calculated as follows:
\begin{align}
\text{MCC} = \frac{\text{TP} \cdot \text{TN} - \text{FP} \cdot \text{FN}}
{\sqrt{(\text{TP}+\text{FP})(\text{TP}+\text{FN})(\text{TN}+\text{FP})(\text{TN}+\text{FN})}}.
\end{align}
In many spatial prediction tasks, such as object detection, it is often unrealistic to expect perfect spatial alignment between predicted and ground-truth positions. To account for this, it can be useful to evaluate classification performance at varying spatial tolerance thresholds, providing a comprehensive view of a model's accuracy beyond a single arbitrary threshold, as done in \cite{accuracy_thres_1, accuracy_thres_2, accuracy_thres_3}. Similarly to how Intersection-over-Union (IoU) thresholds are commonly applied in object detection to determine the acceptability of a predicted bounding box, distance-based thresholds can be used to define whether a prediction is considered a true positive.
By computing MCC at different distance thresholds, practitioners can assess model performance relative to task-specific spatial tolerance. This approach allows for more nuanced evaluation, acknowledging that some use cases may tolerate minor deviations, while others may require high positional precision. The metric is denoted as MCC@ followed by the distance threshold - for example, MCC@2m when a 2-meter threshold is used.

\textbf{Trajectory accuracy} is assessed with two standard metrics in autonomous driving: minADE and minFDE. While minADE captures the minimum average displacement error over the full trajectory, minFDE focuses on the error at the final predicted position.

\subsection{Implementation Details}
In contrast to SceneInformer~\cite{SceneInformer}, we replace pointwise convolution with a fully connected layer, which is mathematically equivalent but yields a slight speed-up in practice, consistent with known PyTorch implementation characteristics~\cite{pytorch}. To accommodate memory limitations on certain GPUs, we reduced the model size. As a result, our model is much more compact, with 2.5M parameters versus 11.3M for SceneInformer. MatchInformer was trained on an NVIDIA GeForce RTX 4090 GPU for 66 hours over 24 epochs, whereas the larger SceneInformer model requires 68 hours to complete only 10 epochs.

\subsection{Experimental Setup}
We benchmark against SceneInformer \cite{SceneInformer}, as it is, to the best of our knowledge, the only existing and reproducible occlusion inference model with open-source code. Following its definition, we test the model at various occlusion levels, which correspond to the probability that nearby agents generate occlusions, with a range from 0\% to 100\%. For a fair comparison, we report both the original results from the paper and our reproduced results using the model that we trained ourselves, based on the authors' provided code. Since SceneInformer did not report MCC, we provide these only for our reproduced version. We also conducted an ablation study on MatchInformer by disabling trajectory prediction. In this configuration, the trajectory loss from Equation~\ref{eq:loss_equation} is deactivated, which allows us to isolate the classification head's contribution. This lets us evaluate how effectively the model can recognize agents on its own, independent of its ability to predict their future trajectories.

\section{Discussion}

\begin{figure*}[thbp]
    \vspace{0.4cm}
    \centering
    \begin{subfigure}{0.98\linewidth}
        \centering
        \raisebox{0.1\height}{\includegraphics[width=0.8\linewidth, trim=0.0cm 0cm 0.0cm 27.5cm, clip]{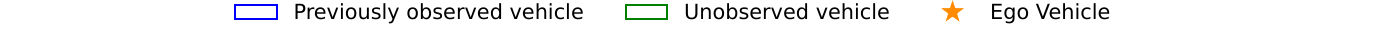}}
    \end{subfigure}
    \begin{subfigure}{0.43\textwidth}
        \centering
        \begin{minipage}{0.48\linewidth}
            \centering
            \includegraphics[width=0.8\linewidth, trim=10cm 11cm 5.5cm 0.3cm, clip]{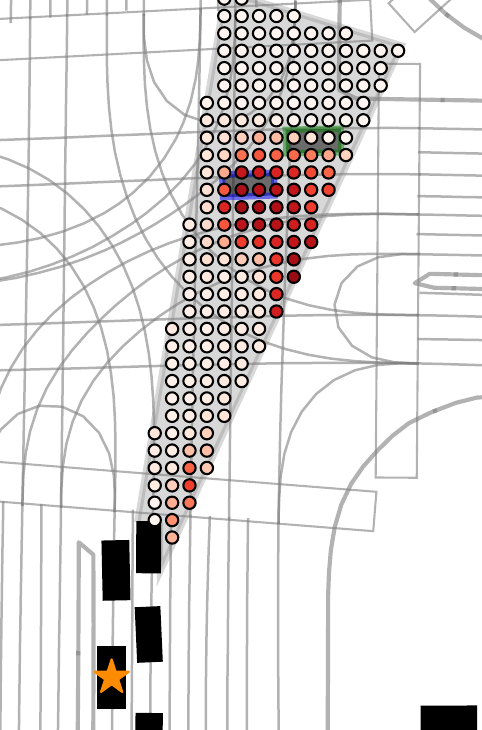}
            \caption*{ SceneInformer (w/o Hungarian Matching)}
        \end{minipage}\hfill
        \begin{minipage}{0.48\linewidth}
            \centering
            \includegraphics[width=0.8\linewidth, trim=10cm 11cm 5.5cm 0.3cm, clip]{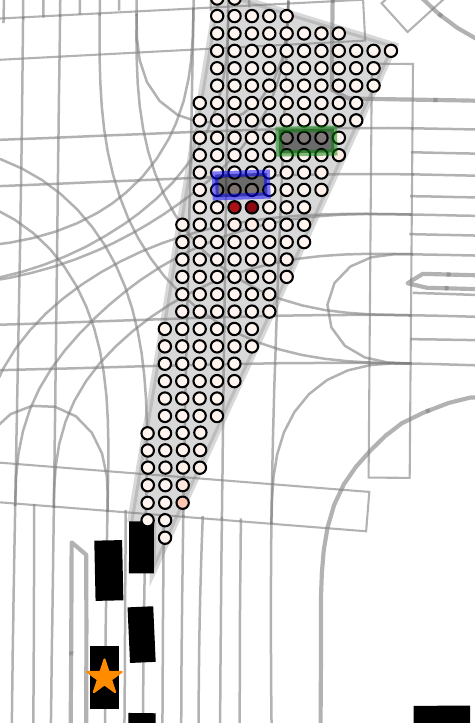}
            \caption*{ \mymodel (with Hungarian Matching)}
        \end{minipage}
        \caption{Scenario 1.}\label{fig:scenario_1} 
    \end{subfigure}\hfill 
    %
    \begin{subfigure}{0.52\textwidth}
        \centering
        \begin{minipage}{0.42\linewidth}
            \centering
            \includegraphics[width=0.8\linewidth, trim=11cm 5cm 1.5cm 2cm, clip]{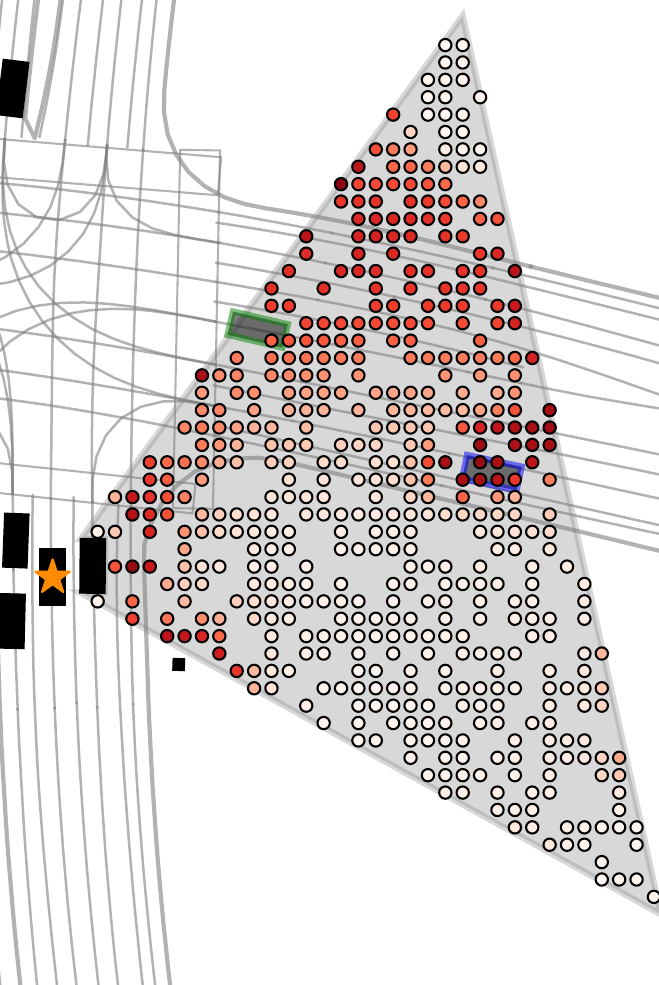}
            \caption*{ SceneInformer (w/o Hungarian Matching)}
        \end{minipage}\hfill
        \begin{minipage}{0.42\linewidth}
            \centering
            \includegraphics[width=0.8\linewidth, trim=11cm 5cm 1.5cm 2cm, clip]{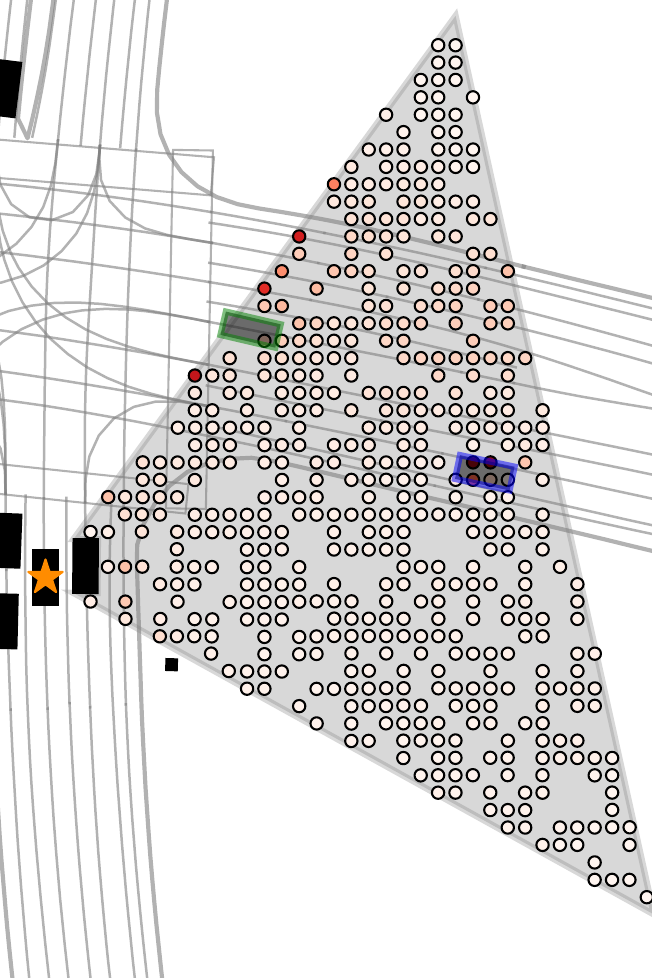}
            \caption*{ \mymodel (with Hungarian Matching)}
        \end{minipage}
        \begin{minipage}{0.12\linewidth}
            \centering
            {\includegraphics[width=0.7\linewidth, trim=20.0cm 7cm 0.0cm 0.1cm, clip]{figures/legend/blue_greenred_bar_legend.pdf}}
        \end{minipage}
        \caption{Scenario 2.}\label{fig:scenario_2}  
    \end{subfigure}

    \caption{Comparison of the occupancy prediction without Hungarian Matching (SceneInformer) and with Hungarian Matching (MatchInformer). In both scenarios, a car (blue) was previously observed when not occluded, and a car (green) was unobserved. In Scenario 1 (a), both models fail to predict the unobserved car. In Scenario 2 (b), SceneInformer predicts occupancy over a large area to actively avoid missing the ground-truth agent, while MatchInformer predicts only a few points within the region, capturing uncertainty without overprediction. This also applies to the previously observed car, where reasoning about speed, heading, and acceleration is required to infer its current position.}
    \label{fig:classification_comparison}
\end{figure*}

Fig.~\ref{fig:classification_comparison} illustrates predictions in two different scenarios. In both scenarios, the ego vehicle has previously observed another vehicle that is now occluded by surrounding traffic. To account for this, the occluded vehicle must be located by estimating its probable current position using its past speed, acceleration, and heading. Interestingly, in both scenarios, SceneInformer predicts a larger spatial area for the previously observed car, while MatchInformer concentrates its predictions on only two or three points with a higher occupancy probability. In \ref{fig:scenario_1}, both SceneInformer and MatchInformer fail to reason for a second occluded agent, which has not been observed in previous timesteps. In contrast, in scenario~\ref{fig:scenario_2}, both models assign a probability of occupancy to the region of the unobserved vehicle. In particular, MatchInformer predicts considerably fewer likely positions in both scenarios, although occasionally at slightly offset locations (see Fig.~\ref{fig:scenario_2}). Furthermore, both models, despite having access to the same historical context, exhibit a significant difference in their reasoning about the right side of the occlusion. While our model effectively leverages this information to correctly infer a low probability of occupancy on the right side—an area previously visible—SceneInformer fails to capitalize on the historical data, often resulting in an incorrect prediction that this specific area is occupied. 
\begin{table}[ht]
\centering
\caption{Comparison of the classification metric MCC ($\uparrow$) at different distance thresholds (0\,m to 4\,m) for occlusion levels ranging from 0\% to 100\%.}
\label{tab:classification}
\resizebox{1.0\columnwidth}{!}{
\begin{tabular}{ccccccc}
\toprule
\multirow{2}{*}{Metric} & \multirow{2}{*}{Model} & \multicolumn{5}{c}{Occlusion Level}  \\
 & & 0\% & 25\% & 50\% & 75\% & 100\%  \\
\midrule
\multirow{3}{*}{MCC@0m} & SceneInf. (reproduced)& 0.20 & 0.18 & 0.16 & 0.14 &  0.13 \\
& MatchInf. (w/o traj)&  \textbf{0.31} &  \textbf{0.27} & \textbf{0.24} & \textbf{0.22} & \textbf{0.20} \\
& MatchInf. (w traj)&  0.27 & 0.24  & 0.22 & 0.20 & 0.18 \\[3pt]
\multirow{3}{*}{MCC@1m} & SceneInf. (reproduced)& 0.20 & 0.18 & 0.16 & 0.14 &  0.13 \\
& MatchInf. (w/o traj) &  \textbf{0.32} &  \textbf{0.28} & \textbf{0.25} & \textbf{0.23} &\textbf{0.21}\\
& MatchInf. (w traj) &  0.28 &  0.26 & 0.22 & 0.20 & 0.19\\[3pt]
\multirow{3}{*}{MCC@2m} & SceneInf. (reproduced)& 0.27 & 0.24 & 0.21 & 0.19 & 0.18\\
& MatchInf. (w/o traj)&  \textbf{0.45} &  \textbf{0.40} & \textbf{0.36} & \textbf{0.33} &\textbf{0.31}\\
& MatchInf. (w traj)&  0.40 &  0.36 & 0.33 & 0.30 & 0.28\\[3pt]
\multirow{3}{*}{MCC@3m} & SceneInf. (reproduced) & 0.30 & 0.26 & 0.23 & 0.22 & 0.20\\
& MatchInf. (w/o traj)&  \textbf{0.47} &  \textbf{0.42} & \textbf{0.38} & \textbf{0.35} &\textbf{0.33}\\
& MatchInf. (w traj)&  0.43 &  0.39 & 0.36 & 0.33 & 0.30\\[3pt]
\multirow{3}{*}{MCC@4m} & SceneInf. (reproduced)  & 0.33 & 0.29 & 0.27 & 0.25 & 0.24 \\
& MatchInf. (w/o traj)&  \textbf{0.48} &  \textbf{0.44} & \textbf{0.40} & \textbf{0.37} &\textbf{0.34}\\
& MatchInf. (w traj)&  0.45 &  0.41 & 0.37 & 0.34 & 0.32 \\
\bottomrule
\end{tabular}}
\end{table}

Table~\ref{tab:classification} presents the MCC values at distance levels from 0\,m to 4\,m for occlusion levels ranging from 0\% to 100\%. Overall, MatchInformer (w/ traj), which includes trajectory prediction, consistently achieves higher classification accuracy than SceneInformer, outperforming it by as much as 58\%. The variant without trajectory prediction, MatchInformer (w/o traj), performs even better, surpassing SceneInformer by up to 74\%. In fact, removing the trajectory loss from MatchInformer's training improves its classification performance by up to 14\%. For both SceneInformer and MatchInformer, the classification metric saturates beyond a distance of 2\,m, indicating that once an agent is detected, its predicted position is typically within 2\,m of the ground-truth agent. These results demonstrate the clear advantage of applying Hungarian Matching to create labels before computing the loss.
\begin{table}[ht]
\centering
\caption{Comparison of the minimum average and final displacement error (minADE $\downarrow$, minFDE $\downarrow$) of the predicted trajectories, reported for occluded/observed agents separately.}
\label{tab:MinADE}
\resizebox{1.0\columnwidth}{!}{
\begin{tabular}{cccccc}
\toprule
\multirow{2}{*}{Model} & \multicolumn{5}{c}{Occlusion Level}  \\
& 0\% & 25\% & 50\% & 75\% & 100\%  \\
\midrule
\multicolumn{6}{c}{minADE $\downarrow$ (occluded/observed)} \\
\midrule
SceneInf. & 0.87/0.26 & 0.91/0.31 & 0.95/0.36 & 0.98/0.40 & 1.00/0.43 \\
SceneInf. (rep.) & 0.85/0.26&0.89/0.30&0.94/0.35&0.98/0.40&0.98/0.43\\
\makecell{MatchInf. \\(w/o yaw)}&0.84/0.25&0.87/0.31&0.90/0.36&0.93/0.40&0.97/0.44\\
\makecell{MatchInf.\\(w yaw)} & \textbf{0.78/0.24} & \textbf{0.81/0.29} & \textbf{0.84/0.33} & \textbf{0.86/0.37} & \textbf{0.88/0.40} \\
\midrule
\multicolumn{6}{c}{minFDE $\downarrow$ (occluded/observed)} \\
\midrule
SceneInf. & 1.43/0.62 & 1.50/0.73 & 1.54/0.83 & 1.59/0.91 & 1.63/0.99 \\
SceneInf. (rep.) & 1.42/0.62 & 1.48/0.73 & 1.53/0.82 & 1.57/0.91 & 1.60/0.98 \\
\makecell{MatchInf.\\(w/o yaw)} & 1.30/0.61 & 1.36/0.72 & 1.40/0.82 & 1.43/0.91 & 1.51/0.98 \\
\makecell{MatchInf.\\(w yaw)} & \textbf{1.17/0.54} & \textbf{1.21/0.64} & \textbf{1.25/0.72} & \textbf{1.28/0.79} & \textbf{1.31/0.86} \\
\bottomrule
\end{tabular}}
\end{table}

Table~\ref{tab:MinADE} reports the minADE and minFDE values. We compare the original reported and reproduced (rep.) SceneInformer results with two MatchInformer variants: one with yaw rate prediction and trajectory adjustment (MatchInf. (w yaw)), and one following the direct trajectory prediction approach of \cite{SceneInformer} (MatchInf. (w/o yaw)). We report metrics separately for occluded and observed agents, computing errors independently for hidden and visible agents at the prediction timestep. As shown in Table \ref{tab:MinADE}, MatchInformer outperforms the baseline in trajectory prediction by at most 12\,\% in minADE and 18\,\% in minFDE, especially for occluded vehicles. Furthermore, the MatchInformer variant without yaw rate prediction and trajectory rotation — i.e., using absolute trajectory prediction — performs worse than the version that incorporates yaw rate and rotation. This underscores the benefit of first estimating the heading angle and then rotating the predicted trajectory accordingly, as it enables the model to focus on capturing different motion modes rather than the precise position of the vehicle within the road layout.
\begin{figure}[h]
    \centering
    \begin{subfigure}{0.98\linewidth}
        \centering
        \includegraphics[scale=0.5, trim=12cm 0.1cm 5cm 25.0cm, clip]{figures/legend/traj_ego_legend.pdf}
    \end{subfigure}
    \begin{subfigure}{0.70\linewidth}  
        \raggedleft
        \includegraphics[width=0.7\linewidth, trim=10cm 11cm 2cm 2.0cm, clip]{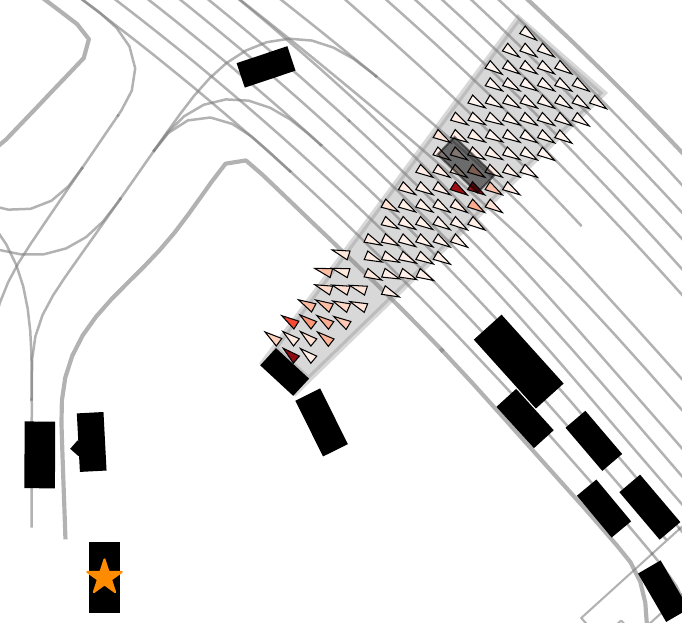}
    \end{subfigure}
    \hfill
    \begin{subfigure}{0.27\linewidth}
        \raggedright
        {\includegraphics[width=0.23\linewidth, trim=20.0cm 7.2cm 0.0cm 0.1cm, clip]{figures/legend/blue_greenred_bar_legend.pdf}}
    \end{subfigure}

    \caption{Visualization of a scenario with the ego vehicle in a parking lot, with its view of the road partially blocked by another parked car. Predicted positions and orientations are shown, with marker tips indicating heading. A key limitation is evident: predicted headings do not rotate 180° for anchor points on the opposite lane, representing oncoming traffic.}
    \label{fig:heading_angle}
\end{figure}
Finally, although our approach performs well both in classification and trajectory prediction, some limitations remain. Fig.~\ref{fig:heading_angle} illustrates a key limitation of MatchInformer: While it correctly captures the orientation of existing vehicles, it does not account for the road layout. Anchor points on the opposite side of the street are predicted with the same orientation instead of rotated 180\,°, as would be required for oncoming traffic. This issue is particularly noticeable when no vehicles are observed on the opposite lane.

    


\section{Conclusion}
In this work, we presented a transformer-based framework that jointly performs multiclass classification and trajectory prediction while explicitly handling occluded agents. By leveraging Hungarian Matching, our method significantly reduces false-positive predictions while maintaining accurate trajectory forecasts. We further refine these forecasts by decoupling the heading from its motion and rotating relative trajectories based on orientation. Experiments demonstrate that our approach outperforms prior methods in both classification and trajectory prediction accuracy. In addition, we introduce the MCC as a robust metric for evaluating imbalanced occupancy predictions, providing a more comprehensive assessment than traditional measures.

Future research will explore a deeper integration of road layout and map information to refine trajectory predictions and better determine an agent's correct heading. In addition, we plan to focus on a sequential planning step that directly leverages our occupancy predictions and their associated uncertainties.



\bibliographystyle{IEEEtran}
\bibliography{references}

\end{document}